# Multi-objective based radiomic feature selection for lesion malignancy classification

Zhiguo Zhou[1], *Member*, *IEEE*, Shulong Li[2], Genggeng Qin[2], Michael Folkert[1], Steve Jiang[1], and Jing Wang[1,*]

*Abstract*—*Objective:* **Accurately classifying the malignancy of lesions detected in a screening scan is critical for reducing false positives. Radiomics holds great potential to differentiate malignant from benign tumors by extracting and analyzing a large number of quantitative image features. Since not all radiomic features contribute to an effective classifying model, selecting an optimal feature subset is critical.** *Methods:* **This work proposes a new multi-objective based feature selection (MO-FS) algorithm that considers sensitivity and specificity simultaneously as the objective functions during feature selection. For MO-FS, we developed a modified entropy based termination criterion (METC) that stops the algorithm automatically rather than relying on a preset number of generations. We also designed a solution selection methodology for multi-objective learning that uses the evidential reasoning approach (SMOLER) to automatically select the optimal solution from the Pareto-optimal set. Furthermore, we developed an adaptive mutation operation to generate the mutation probability in MO-FS automatically.** *Results:* **We evaluated the MO-FS for classifying lung nodule malignancy in low-dose CT and breast lesion malignancy in digital breast tomosynthesis.** *Conclusion:* **The experimental results demonstrated that the feature set selected by MO-FS achieved better classification performance than features selected by other commonly used methods.** *Significance:* **The proposed method is general and more effective radiomic feature selection strategy.**

## I. INTRODUCTION

ACCURATELY classifying the malignancy of lesions detected in a screening test is critical for reducing false positives and unnecessary follow-up tests. Several studies have shown that computer-aided diagnosis schemes can efficiently assist radiologists in differentiating malignant from benign tumors [1-3]. In recent years, radiomics has shown great potential for classifying lesion malignancy by extracting and analyzing a large number of quantitative imaging features [4, 5]. For example, radiomics has been successfully applied to lung nodule classification [6, 7] and breast lesion malignancy classification [8, 9]. In radiomics, typically several hundred features are extracted from segmented lesions [4]. However, not all of these features are discriminative, and many are correlated, redundant, or even irrelevant, which may reduce the model's performance. In addition, a high-dimensional feature space increases the model's complexity and may cause over-fitting.

Selecting a subset of relevant features from the original feature set (i.e., feature selection) is a critical step in radiomics model construction, as it can simplify the predictive model, increase the model's performance, reduce the dimensionality of the feature space, and speed up the learning processing [10]. Current feature selection algorithms can be divided into two categories: filter approaches and wrapper approaches. Filter approaches use a suitable ranking criterion to score features and remove those that fall below a certain threshold [11]. Several filter approaches have been used for radiomic feature selection, including correlation coefficient analysis (CCA), mutual information maximization (MIM), minimum redundancy maximum relevance (mRMR), and relevance in estimating features (RELIEF) [12]. Filter approaches can be used as an initial step to remove redundant features, while wrapper approaches can further select features by evaluating the predictive performance of classifiers using the selected feature set. Sequential forward selection (SFS) [11]and sequential backward selection (SBS) [11] are the two classic wrapper algorithms. The evolutionary computation (EC) based feature selection method has also gained much attention and shown some success in recent years [10].

Because predictive model performance is evaluated in wrapper based feature selection approaches, evaluation criteria play an important role in selecting an appropriate feature set. Accuracy or area under the curve (AUC) has been used widely for evaluating model performance. However, a single metric may not suffice, especially for imbalanced positive and negative cases, as both sensitivity and specificity are required for a diagnostic procedure or modality [13-15]. Therefore, in this work, we consider feature selection as a multi-objective problem and use evolutionary computation for multi-objective optimization. Several multi-objective evolutionary algorithm based feature selection methods have been proposed, including genetic algorithm based methods [16, 17], particle

This work was supported in part by the American Cancer Society (ACS-IRG-02-196) and the US National Institutes of Health (5P30CA142543). Corresponding author: Jing Wang, e-mail: Jing. Wang@UTSouthwestern.edu.

Zhiguo Zhou, Steve Jiang and Jing Wang are with Medical Artificial Intelligence and Automation Laboratory (MAIA Lab), UT Southwestern Medical Center, Dallas, TX, USA.

Zhiguo Zhou, Michael Folkert, Steve Jiang and Jing Wang are with Department of Radiation Oncology, UT Southwestern Medical Center, Dallas, TX, USA.

Shulong Li and Genggeng Qin are with the school of Biomedical Engineering, Southern Medical University, Guangzhou, China.

swarm optimization based methods [18, 19], and a colony optimization based method [20]. However, these algorithms do not consider two issues. First, the number of generations that the multi-objective evolutionary algorithm runs is fixed arbitrarily. If this number is not large enough, we may only get local optimal solutions, and it may get stuck in a part of the Pareto-optimal solution set [21]. Second, the optimal solution is selected manually from the Pareto-optimal solution set. Since the Pareto-optimal solution set always contains too many solutions, it is difficult for the decision maker to select the preferred solution [22].

To overcome these issues, we propose a new multi-objective based feature selection (MO-FS) algorithm in this work. The improvements of MO-FS include: 1) a modified entropy based termination criterion (METC) that stops the algorithm automatically rather than relying on a preset fixed number of generations; 2) a selection methodology for multi-objective learning using the evidential reasoning approach (SMOLER) that selects the optimal solution from the Pareto-optimal set automatically; and 3) an adaptive mutation operation designed to calculate mutation probability automatically instead of using a manually preset mutation probability.

In the proposed MO-FS, METC was developed based on an entropy based termination criterion for multi-objective evolutionary computation [21]. In addition to measuring the dissimilarities between the objective functions through relative entropy, as measured in the entropy based termination criterion, METC measures the dissimilarities for the selected feature set to select the most stable radiomic features. In SMOLER, the optimal solution selection rules are designed first and then combined using the evidential reasoning approach, which was originally proposed to deal with multiple attribute decision analysis problems [23-25]. In the adaptive mutation operation, the mutation probability is determined by correlation coefficients among individuals in one solution so that non-redundant features are selected with high probability. We evaluated the performance of MO-FS in two datasets: lung nodule malignancy classification in computed tomography (CT) [26] and breast lesion malignancy classification in digital breast tomosynthesis.

## II. METHODS

### A. General description

The MO-FS framework for feature selection is shown in Fig. 1. Before beginning the procedure, features are extracted from segmented images. MO-FS consists of two phases: (1) generating the Pareto-optimal solution set; and (2) selecting the best solution through SMOLER. Then, the discriminative features are selected.

When generating the Pareto-optimal solution set in the first phase, sensitivity and specificity are considered simultaneously as the objective functions. This phase consists

of six steps, as follows:

Step 1: Initialization. All solutions in the population are generated randomly and are denoted by $S(j) = \{s_1, \cdots, s_P\}, j = 0$, where $P$ is the population size and j is the generation number. Each solution is denoted by $s_p = \{I_p^1, \cdots, I_p^N\}$, where $N$ is the number of individuals (features). Binary coding is adopted for each solution in the population. In detail, "1" means the selected individual, while "0" means the unselected individual.

Step 2: Clonal operation. The proportional cloning strategy [27] is used, and the obtained cloned population $C(j)$ is generated.

Step 3: Adaptive mutation operation. The mutation probability $MP$ is generated automatically in this step, and the mutated solution set $M(j)$ is generated based on $C(j)$. Then, the new solution set $F(j)$, which combines $S(j)$ and $M(j)$, is generated.

Step 4: Deleting operation. When F(j) yields multiple solutions with the same sensitivity and specificity, only the solution with the highest AUC is kept. The remaining solutions after the deletion operation constitute the new solution set $DF(j)$. When $size(DF(j)) < P$, the algorithm returns to step 2; otherwise, it continues to step 5.

Step 5: Updating solution set. AUC based non-dominated sorting [13] is used to update the solution set, and $UD(j)$ is generated.

Step 6: Termination detection. If the solution set $UD(j)$ satisfies the METC, phase 1 ends; otherwise, let $j = j + 1$, $S(j) = UD(j)$, and the algorithm returns to step 2.

The second phase is divided into three steps as follows:

Step 1: Extracting Pareto front set ($PD$) is from $UD(j)$.

Step 2: Selecting best solution $D^*$ from $PD$ through SMOLER.

Step 3: Obtaining the corresponding selected feature set $SF$.

The following sections describe the adaptive mutation operation, modified entropy based termination criterion, and SMOLER in detail.

### B. Adaptive mutation operation

In the adaptive mutation operation, the mutation probability is set automatically based on correlation coefficients among features. First, the correlation coefficient matrix $R$ is calculated as [28]:

$$R = \begin{bmatrix} |r_{1,1}| & \cdots & |r_{1,n}| \\ \vdots & \ddots & \vdots \\ |r_{n,1}| & \cdots & |r_{n,n}| \end{bmatrix}, \tag{1}$$

where $r_{i,j}$ represents the correlation coefficient between two features. Assume that there are $K$ selected features in a solution $s_p$. Then, correlation coefficients denoted by $\{r_{i,1}, \cdots, r_{i,K}\}$ between any individual feature $I_p^i$ and $K$ selected features can be extracted from $R$. If $I_p^i$ is a selected feature, then the mutation probability of this feature is calculated as:

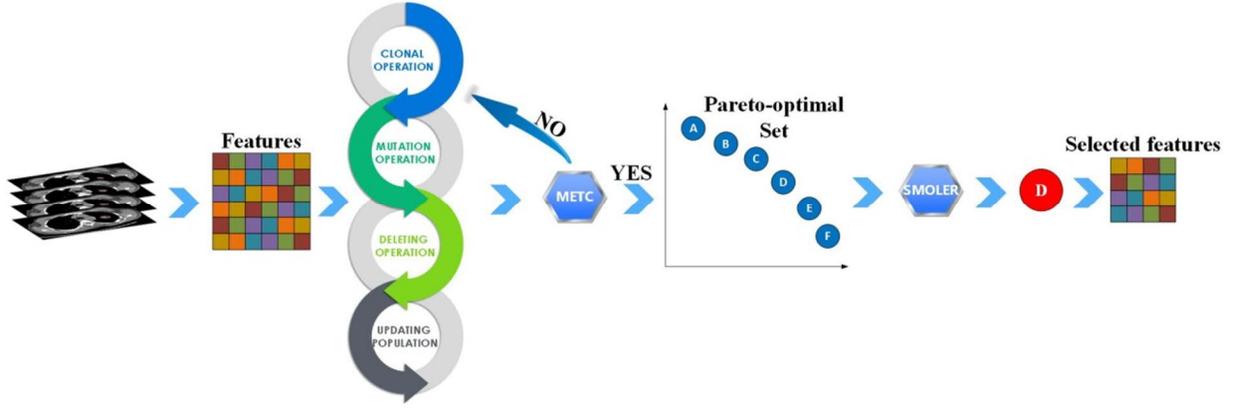

Fig. 1. MO-FS Framework for feature selection.

$$MP_i = \frac{\sum_{k=1, k \neq i}^{K} r_{i,k}}{K},\qquad(2)$$

If $I_p^i$ is not selected in $s_p$, then,

$$MP_i = 1 - \frac{\sum_{k=1, k \neq i}^{K} r_{i,k}}{K}.\qquad(3)$$

The above formulation for calculating mutation probability $MP_i$ for feature $I_p^i$ is based on the idea that, if the selected feature $I_p^i$ is highly correlated with other selected features, then a higher mutation probability is calculated according to equation (2), so this feature has a higher chance to mutate and is unselected. On the other hand, if an unselected feature is highly correlated with other selected features, then a lower mutation probability for this unselected feature is calculated according to equation (3), so this feature has a lower chance to mutate and remains unselected. A random mutation $RP_i$ is generated. If $MP_i > RP_i$, the mutation will perform; otherwise, the algorithm continues to the next individual.

### C. Modified entropy based termination criterion

Same as entropy based termination criterion [21], our modified entropy based termination criterion (METC) also consists of four stages (see Table I): (1) generating a cell identification number for each solution; (2) obtaining the probability distribution for objective functions; (3) measuring the dissimilarity; and (4) detecting termination.

In the first stage, the solutions in the population are mapped into the range [0, 1] through:

$$\overline{s_j} = \frac{s_j - O_{min,j}}{O_{max,j} - O_{min,j}}, for\ j = 1, \cdots, M,\qquad(4)$$

where $O_{min,j}$ and $O_{max,j}$ define the minimum and maximum values, respectively, for the $jth$ objective, and $M$ is the objective number. Let $n_b$ denote a fixed number of intervals, which is defined as an anchor point and bin width for each dimension. Assume that a vector $B = \{(\frac{0}{n_b}), (\frac{1}{n_b}), \dots, (\frac{n_b}{n_b})\}$ with size $n_b + 1$ defines a set of intervals, which satisfies $B_{k_j} \leq \overline{s_j} \leq B_{k_j+1}, k_j \in [0, \cdots, n_b - 1]$. The cell identification number can be calculated as:

$$c = \sum_{j=1}^{M} k_j \times n_b^{j-1}.\qquad(5)$$

Based on the cell identification number, the multi-dimensional histogram is generated in the second stage. Assume that there are $N = \{N_1, \cdots, N_{n_b \times n_b}\}$ numbers in each bin. The probability distribution $P$ for the current population is

calculated as:

$$p_i = \frac{N_i}{N}, i = 1, \cdots, n_b \times n_b,\qquad(6)$$

where $\widehat{N}$ represents the population size. Assume that $Q = \{q_1, \cdots, q_{n_b \times n_b}\}$ is the probability distribution in the next generation.

Meanwhile, it is assumed that there are $N = \{N_1, \cdots, N_{n_b \times n_b}\}$ ($n_b$ is the number of intervals) numbers in each bin for population $P$. For each bin $B_i, i = 1, \cdots, n_b \times n_b$, the selected features in each solution are denoted by $F_i^{B_i, P} = \{F_1^{B_i, P}, \cdots, F_{N_i}^{B_i, P}\}, i = 1, \cdots, n_b \times n_b$, where $F_i^{B_i, P}$ is a binary vector. Hence, the probability distribution for selected features in each bin is calculated as:

$$p_{B_i}^P = \frac{\sum_{j=1}^{N_i} F_j^{B_i, P}}{N_i \times N}, i = 1, \cdots, n_b \times n_b,\qquad(7)$$

where $N$ is the feature number and $j$ represents the number in each bin. Similarly, we can obtain $p_{B_i}^Q$ for the next generation $Q$. Similarly, we can obtain $p_{B_i}^Q$ for the next generation $Q$.

The dissimilarity of objective functions between $P$ and $Q$ can be calculated in the third stage based on different situations. For intersection set ($p_i \neq 0$ and $q_i \neq 0$):

$$D(p, q)_I = KL(p|q) + KL(q|p),\qquad(8)$$

where,

$$KL(p|q) = -\sum \frac{p(x_i)}{2} log \left\{ \frac{q(x_i)}{p(x_i)} \right\},$$
$$KL(q|p) = -\sum \frac{q(x_i)}{2} log \left\{ \frac{p(x_i)}{q(x_i)} \right\}.\qquad(9)$$

For non-interaction set ($p_i = 0$ or $q_i = 0$):

$$D(p, q)_y = D(p, q)_{y_p} + D(p, q)_{y_q},\qquad(10)$$

where,

$$D(p, q)_{y_p} = -\sum \frac{p(x_i)}{2} log\{p(x_i)\},$$
$$D(p, q)_{y_q} = -\sum \frac{q(x_i)}{2} log\{q(x_i)\}.\qquad(11)$$

Therefore, the dissimilarity of objective functions $D_O(p, q)$ between generations is:

$$D_O(p, q) = D(p, q)_I + D(p, q)_y.\qquad(12)$$

Similarly, the dissimilarity of selected features $D_{B_i}(p, q)$ for each bin between $P$ and $Q$ can be calculated through equations (8)-(12), and the final dissimilarity $D(p, q)$ is:

$$D(p, q) = D_O(p, q) + \sum_{i=1}^{n_b \times n_b} D_{B_i}(p, q).\qquad(13)$$

In the final stage, the generation counter is denoted by $t$,

and the current generation is denoted by $i$. $D_i$ represents the $D_O(p, q)$ in the $i$th generation, and $M_t$ and $S_t$ are the mean and standard deviation of $D_O(p, q)$ from the first to the $i$th generation, calculated as:

$$M_1 = D_1 \text{ and } M_t = \frac{1}{t}\sum_{i=1}^{t} D_i, \text{where } t \geq 2, \quad (14)$$

$$S_t = \frac{1}{t}\sum_{i=1}^{t}(D_i - M_t)^2. \quad (15)$$

When $M_t$ and $S_t$ in a manually defined number ($n_s$) of successive generations coincide up to a pre-specified number of decimal ($n_p$), the algorithm will be terminated.

When the proposed MO-FS algorithm achieves the global optimization in an ideal situation, the selected features should be fixed after a certain number of generations. At that time, even though we run the algorithm for the next generation, we should obtain the same selected features. In other words, the dissimilarity of selected features between these two successive generations is 0. Therefore, we may obtain more stable selected features when we add the dissimilarity of selected features into the final dissimilarity measure.

TABLE I
BRIEF DESCRIPTION OF METC

**Input:** Multiple successive generations
**Step 1:** Generating cell identification number. The unique cell identification number for all solutions in each generation is calculated through Eq. (5).
**Step 2:** Obtaining probability distribution. The probability distribution for the objective functions is calculated through Eq. (6), while the probability distribution for selected features in each bin is calculated through Eq. (7).
**Step 3:** Measuring dissimilarity. $D_O(p, q)$ and $D_{B_i}(p, q)$ are calculated first, then $D(p, q)$ is calculated through Eq. (13).
**Output:** Detecting termination. When $M_t$ and $S_t$ coincide up to a pre-specified number of decimal places, the algorithm terminates.

### D. Solution selection methodology for multi-objective learning algorithm using the evidential reasoning approach (SMOLER)

The procedures of SMOLER are shown in Table II. Assume that there are $M_1$ objective functions and $K$ solutions $P = \{P_1, P_2, \cdots, P_K\}$ in the Pareto solution set. To select the optimal solution, several decision rules need to be set. These rules include two types: one type is based on the objective function denoted by $F_1$, and the other type is based on the preference and prior knowledge denoted by $F_2$. Therefore, two objective functions (sensitivity and specificity) are taken as the first two rules, and the $F_1$ is,

$$F_1^k = \{f_{sen}^k, f_{spe}^k\}, k = 1,2,\cdots, K. \quad (16)$$

Area under the curve (AUC), an important evaluation criterion to determine whether the model is reliable, is adopted as the first rule in $F_2$. In addition, to select a solution with balanced sensitivity and specificity, the relative distance is defined to evaluate the solution, that is,

$$f_{RD}^k = \left| f_{sen}^k - f_{spe}^k \right|^2, k = 1,2,\cdots, K. \quad (17)$$

So $F_2$ is,

$$F_2^k = \{f_{AUC}^k, f_{RD}^k\}, k = 1,2,\cdots, K. \quad (18)$$

When combining $F_1$ and $F_2$, the final rule set $F$ is,

$$F = \{f_1^k, f_2^k, f_3^k, f_4^k\}, k = 1,2,\cdots, K, \quad (19)$$

where $f_1^k = f_{sen}^k$, $f_2^k = f_{spe}^k$, $f_3^k = f_{AUC}^k$, and $f_4^k = f_{RD}^k$. Assume that there are $N$ reference points for each rule in F, which are used to assess each solution for all rules [29], and the reference points are denoted by $H = \{H_1, H_2, \cdots, H_N\}$. For the first three rules in $F$, the higher the values, the better the solutions. For the fourth rule, the solutions are better when the value is lower. Thus, the reference value $H_{i,j}$ for each rule i at reference point $j$ is calculated as follows:

$$H_{i,j} = \begin{cases} min(f_i^k) + (j-1) \times \frac{max(f_i^k) - min(f_i^k)}{N-1}, i = 1,2,3 \\ max(f_i^k) - (j-1) \times \frac{max(f_i^k) - min(f_i^k)}{N-1}, i = 4 \end{cases}, \quad (20)$$

where $j = 1,2,\cdots, N, k = 1,2,\cdots, K$. The optimal solution selection can be modeled using the following expectations [29]:

$$S(P_k) = \{H_{i,j}, \beta_{i,j}(P_k), j = 1,2,\cdots, N\},$$
$$i = 1,2,\cdots, M, k = 1,2,\cdots, K, \quad (21)$$

where $\beta_{i,j}(P_k) \geq 0$ and $\sum_{j=1}^{N} \beta_{i,j}(P_k) = 1$. $\beta_{i,j}(P_k)$ represents a degree of belief for solution $P_k$. Similar to the reference value $H_{i,j}$, $\beta_{i,j}(P_k)$ is calculated under the two situations (rules 1-3 and rule 4) in the third step. For rules 1-3, $\beta_{i,j}(P_k)$ is calculated as:

$$\beta_{i,j}(P_k) = \frac{H_{i,j+1} - f_i^k}{H_{i,j+1} - H_{i,j}}, \quad \beta_{i,j+1}(P_k) = 1 - \beta_{i,j}(P_k),$$
$$when\ H_{i,j} \leq f_i^k \leq H_{i,j+1}, \beta_{i,p}(P_k) = 0 \quad p = 1,2,\cdots, N,$$
$$and\ p \neq j, j+1, i = 1,2,3. \quad (22)$$

For the fourth rule, $\beta_{i,j}(P_k)$ is calculated as:

$$\beta_{i,j}(P_k) = \frac{H_{i,j} - f_i^k}{H_{i,j} - H_{i,j+1}}, \beta_{i,j+1}(P_k) =$$
$$1 - \beta_{i,j}(P_k), when\ H_{i,j+1} \leq f_i^k \leq H_{i,j},$$
$$\beta_{i,p}(P_k) = 0 \quad p = 1,2,\cdots, N, and\ p \neq j, j+1, i = 4. \quad (23)$$

The belief degrees for each rule can generate a belief degree matrix for all feasible solutions:

$$S_k = \begin{bmatrix} \beta_{1,1} & \beta_{1,2} & \cdots & \beta_{1,N} \\ \beta_{2,1} & \beta_{2,2} & \cdots & \beta_{2,N} \\ \beta_{3,1} & \beta_{3,2} & \cdots & \beta_{3,N} \\ \beta_{4,1} & \beta_{4,2} & \cdots & \beta_{4,N} \end{bmatrix}, k = 1,2,\cdots, K. \quad (24)$$

In the fourth step, all the rules in $S_k$ are combined through the ER approach. Assume that the weights for each rule are denoted by $\omega_i, i = 1, \cdots, 4$, which satisfies the following constraints:

$$0 \leq \omega_i \leq 1, \sum_{i=1}^{M} \omega_i = 1. \quad (25)$$

The final assessment $D(P_k)$ for solution $P_k$ is represented by:

$$D(P_k) = \{(H_{k,j}, \beta_{k,j}), j = 1,2,\cdots, N\}, k = 1,2,\cdots, K. \quad (26)$$

Then, the belief degree $\beta_{k,j}$ for solution $P_k$ at each reference point $j$ in $D(P_k)$ is calculated using the evidential reasoning algorithm [30]:

$$\beta_{k,j} = \frac{\mu \times \left[\prod_{i=1}^{M}(\omega_i \beta_{i,j}(P_k) + 1 - \omega_i \sum_{j=1}^{N} \beta_{i,j}(P_k)) - \prod_{i=1}^{M}(1 - \omega_i \sum_{j=1}^{N} \beta_{i,j}(P_k))\right]}{1 - \mu \times \left[\prod_{i=1}^{M}(1 - \omega_i)\right]}, \quad (27)$$

$$\mu = \left[\sum_{j=1}^{N} \prod_{i=1}^{M}(\omega_i \beta_{i,j}(P_k) + 1 - \omega_i \sum_{j=1}^{N} \beta_{i,j}(P_k)) - (N-1)\prod_{i=1}^{M}\left(1 - \omega_i \sum_{j=1}^{N} \beta_{i,j}(P_k)\right)\right]^{-1}. \quad (28)$$

To select the optimal solution, the utility for $P_k, k = 1,2,\cdots, K$ is then calculated. Since there are $N$ reference points, $N$ evaluation grades are also needed. Assume that the utility of the grades $u(H_j)$ is equidistantly distributed in the utility space, i.e., $u(H_j) = \frac{j-1}{N-1}, j = 1,2,\cdots, N$. Then, the utility for $P_k$ is calculated:

$$U(P_k) = \sum_{j=1}^{N} u_j \beta_{k,j}, k = 1, 2, \cdots, K. \quad (29)$$

The final solution $P^*$ is selected by:

$$U^* = max(U(P_k), k = 1, 2, \cdots, K). \quad (30)$$

TABLE II
BRIEF DESCRIPTION OF SMOLER

**Input:** Pareto solution $P = \{P_1, P_2, \cdots, P_K\}$, weight $\omega_i, i = 1, 2, \cdots, M$, and number of reference points $N$.

**Step 1:** Generate solution selection rules $F^k, k = 1, 2, \cdots, K$.

**Step 2:** Calculate reference values $H_{i,j}, i = 1, 2, \cdots, M, j = 1, 2, \cdots, N$.

**Step 3:** Calculate belief degrees $\beta_{i,j}(P_k), i = 1, 2, \cdots, M, j = 1, 2, \cdots, N, k = 1, 2, \cdots, K$.

**Step 4:** Calculate utilities $U(P_k), k = 1, 2, \cdots, K$.

**Output:** Select the final solution $P^*$.

## III. EXPERIMENTS AND ANALYSIS

### A. Datasets

We first evaluated MO-FS using the Lung Image Database Consortium and Image Database Resource Initiative (LIDC-IDRI) dataset, which consists of 1,010 patients with thoracic computed tomography (CT) imaging and annotation results from four radiologists. In this dataset, 7,371 lesions were marked as nodules by at least one radiologist, and malignancy suspiciousness was rated on a scale of 1 to 5 (1 indicates the lowest malignancy suspiciousness, and 5 indicates the highest). This study considered nodules 3 mm or larger in size. We obtained the malignancy suspiciousness rate by averaging the suspicion level from the radiologists. After removing ambiguous nodules with an average suspicion level of 3, 431 malignant and 795 benign nodules remained. All nodules were contoured manually by radiologists. Typical malignant and benign nodules in this dataset are shown in Fig. 2.

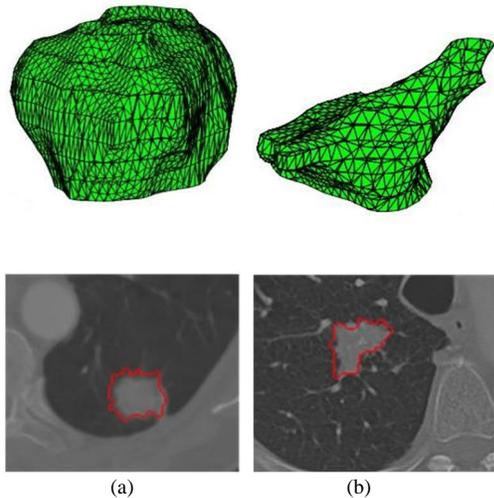

Fig. 2. Benign (a) and malignant (b) lung nodules. The first row is the 3D tumor, and the second row is the corresponding 2D CT images.

We then evaluated MO-FS by classifying breast lesion malignance in a dataset of digital breast tomosynthesis (BLM-DBT). The patient DBT images comprise 278 malignant and 685 benign cases. Each lesion on DBT was initially contoured by one of eight radiologists with more than 3 years of experience in breast cancer diagnosis. Two additional radiologists with more than 5 years of experience reviewed and modified the contours if needed. The malignancy status was validated through biopsy. Typical malignant and benign tumors are shown in Fig. 3.

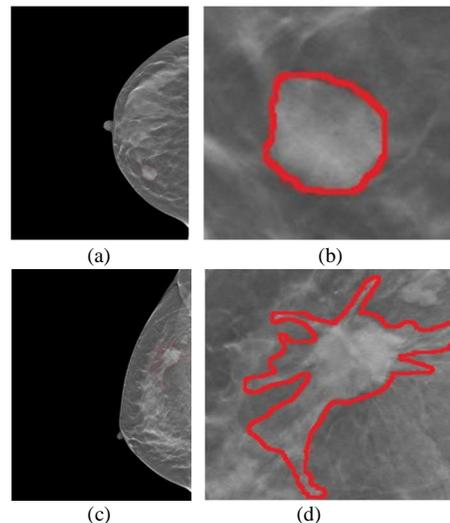

Fig. 3. Benign and malignant breast tumors. (a) is the original image and (b) is the tumor region for a benign case. (c) is the original image and (d) is the tumor region for a malignant case.

### B. Setup

Three types of radiomic features – intensity, texture, and geometry – and 257 features in total were extracted. According to the previous studies on the classical multi-objective evolutionary algorithms [31, 32], setting the population size 100 can cover the most Pareto solutions and obtain the promising results. We follow these recommendations, and choose 100 as the population size. In METC, the number of intervals $n_b$ was set as 4, while both the number ($n_s$) of successive generations and the number of decimal ($n_p$) were set as 2. The reason we set $n_b$ as 4 is because generating 25 cells is reasonable for calculating the probability distribution based on the feature number and population size. If the population size and feature number is larger for other datasets, $n_b$ should be set as a larger value accordingly. On the other hand, $n_s$ and $n_p$ is dependent on the computational complexity of the problem. To determine which values are more reasonable, the analysis based on the experiments was performed, and the results showed that when $n_s$ and $n_p$ were set as 2, we have already obtained satisfactory results. However, when the problem or dataset is more complex, the large values may be needed. In SMOLER, the reference point number was set as 5. Our work on SMOLER is inspired by the work in [29]. In this work, the reference point number was set 5 and a promising result can be achieved. So we also chose 5 as the reference point number. Among the four rules, sensitivity and specificity are the main objective functions, so they are slightly more important than relative distance and AUC. Accordingly, the weight was set as $\omega = \{0.3, 0.3, 0.2, 0.2\}$. To demonstrate MO-FS's performance, we compared it with five commonly used feature selection methods: correlation based feature selection (CFS) [33], evolutionary computation based feature selection (ECFS) [34], minimum redundancy maximum relevance (mRMR), relevance in estimating features (RELIEF), sequential forward selection (SFS) with AUC as the objective functions, and sequential forward floating selection (SFFS). Since F1-score also consider false positive and false negative simultaneously, SFS with F1 score (SFS-F1) are also compared. We also compared with the model with all the input features (ALL). We used support vector machine with radial basis function as a training model for all feature selection methods and performed two-fold cross-validation. Since our aim is to select the optimal features, the gamma parameter in SVM is set as constant. In our cross validation, we divided the data based on the patients in LIDC-IDRI dataset, i.e., if several lesions are from one patient, we will divide all of them into training or testing dataset. Since there is only one tumor for each case in BLM-DBT dataset, we divide the dataset based on the patient case. In both datasets, one patient only has one associated image. Therefore, there is no overlap between case studies in training and testing sets. A trained model evaluated the performance through the selected feature set. Area under the curve (AUC), accuracy (ACC), sensitivity (SEN), and specificity (SPE) were used for quantitative evaluation. All methods were performed ten times. The mean and standard deviation for each evaluation criterion were calculated from 10 times results.

*C. Results and analysis*

Table III summarizes the model performance on the LIDC-IDRI dataset after being trained using the feature sets selected by six feature selection methods. The MO-FS obtained better performance than the other methods. MO-FS also obtained the smallest difference (0.0481) between sensitivity and specificity. Similar results were obtained in BLM-DBT as shown in Table IV. When comparing with ALL, the analysis on two studies demonstrated that when selecting effective features, better results can be obtained. When using F1-score as the objective function, it shows that SFS can obtain better sensitivity. Although the specificity between SFFS and MO-FS is similar, the sensitivity of SFFS is still lower than MO-FS. The sensitivity and specificity are more balanced as well in MO-FS. In both studies, MO-FS achieves the highest AUC and accuracy due to the reliability of the selected feature set.

TABLE III
FEATURE SELECTION PERFORMANCE FOR DIFFERENT METHODS IN LIDC-IDRI.
THE BEST RESULTS ARE IN BOLD

| Method | AUC | ACC | SEN | SPE |
|---|---|---|---|---|
| ALL | 0.896±0.001 | 0.847±0.002 | 0.792±0.002 | 0.875±0.002 |
| CFS | 0.899±0.000 | 0.847±0.001 | 0.802±0.002 | 0.871±0.002 |
| ECFS | 0.910±0.000 | 0.862±0.001 | 0.805±0.002 | 0.893±0.002 |
| mRMR | 0.906±0.000 | 0.851±0.001 | 0.804±0.004 | 0.877±0.001 |
| RELIEF | 0.913±0.000 | 0.865±0.002 | 0.804±0.002 | 0.899±0.002 |
| SFS | 0.906±0.001 | 0.865±0.003 | 0.753±0.002 | 0.906±0.002 |
| SFS-F1 | 0.900±0.007 | 0.870±0.002 | 0.764±0.007 | 0.905±0.005 |
| SFFS | 0.915±0.006 | 0.873±0.003 | 0.770±0.009 | 0.905±0.005 |
| MO-linear | 0.908±0.001 | 0.856±0.002 | 0.812±0.001 | 0.889±0.006 |
| MO-FS | **0.935±0.001** | **0.889±0.001** | **0.858±0.005** | **0.907±0.004** |

TABLE IV
FEATURE SELECTION PERFORMANCE FOR DIFFERENT METHODS IN BLM-DBT.
THE BEST RESULTS ARE IN BOLD

| Method | AUC | ACC | SEN | SPE |
|---|---|---|---|---|
| ALL | 0.677±0.003 | 0.635±0.002 | 0.623±0.005 | 0.665±0.003 |
| CFS | 0.732±0.000 | 0.665±0.001 | 0.669±0.016 | 0.658±0.013 |
| ECFS | 0.683±0.006 | 0.642±0.006 | 0.635±0.014 | 0.660±0.016 |
| mRMR | 0.759±0.004 | 0.718±0.004 | 0.644±0.008 | 0.747±0.004 |
| RELIEF | 0.776±0.002 | 0.737±0.005 | 0.650±0.007 | 0.772±0.007 |
| SFS | 0.777±0.003 | 0.741±0.003 | 0.571±0.014 | 0.809±0.006 |
| SFS-F1 | 0.782±0.002 | 0.745±0.004 | 0.576±0.009 | 0.814±0.001 |
| SFFS | 0.808±0.003 | 0.718±0.003 | 0.674±0.005 | **0.815±0.003** |
| MO-linear | 0.771±0.002 | 0.736±0.003 | 0.742±0.005 | 0.762±0.002 |
| MO-FS | **0.819±0.004** | **0.756±0.005** | **0.764±0.002** | 0.815±0.002 |

We also show the confusion matrix including true positive (TP), false positive (FP), true negative (TN) and false negative (FN) [35] for eight feature selection methods in Fig. 4. Compared with other methods, MO-FS obtains higher TP and TN, and lower FP and FN.

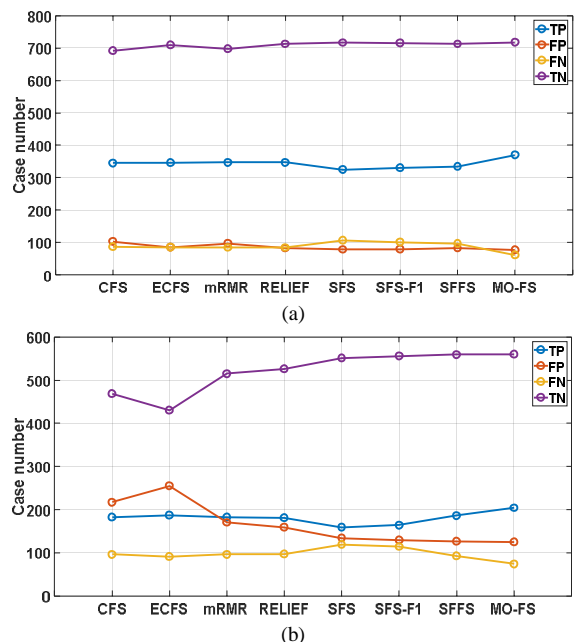

Fig. 4. Confusion matrix results for eight feature selection methods. (a) LIDC-IDRI and (b) BLM-DBT.

We also compared the performance after using our MO-FS feature selection with the deep learning method. Since AlexNet [36] is a classical deep learning method, and already achieved great performance in many fields, it is chosen in this study. The comparative results on two dataset are shown in Fig. 5 and results demonstrate that MO-FS outperforms AlexNet based on different evaluation criteria.

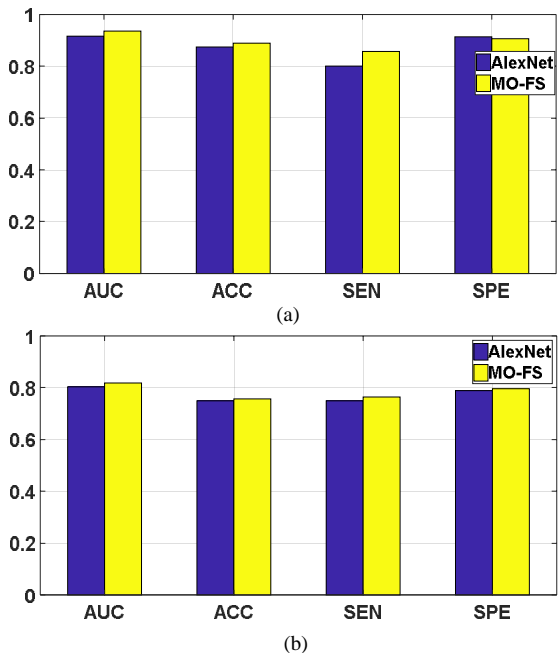

Fig. 5. Comparative results with deep learning. (a) LIDC-IDRI and (b) BLM-DBT.

Fig. 6 shows the selected features for two datasets through MO-FS where each bar indicates that the corresponding feature is selected. If two bars (green and blue) for one feature are shown in both two datasets, it means that this feature is selected by the two studies. There are 92 and 87 selected features for LIDC-IDRI and BLM-DBT, respectively. It is shown that two datasets select the different feature subset while some features are overlapped. This is because the two studies aim at two different clinical problems for different disease sites, and the feature combinations is expected to be different. Among three types of radiomic features, texture features are selected most for both studies. Indeed, several studies have demonstrated that texture features play an important role for lung nodule [37] and breast lesion classification [38]. On the other hand, BLM-DBT picks more geometry features. This is because tumor shape has become one of the most indicators for classifying breast lesion malignancy. Most benign masses are compact, roughly elliptical and well-circumscribed, while malignant lesions always have an irregular appearance and a blurred boundary [39].

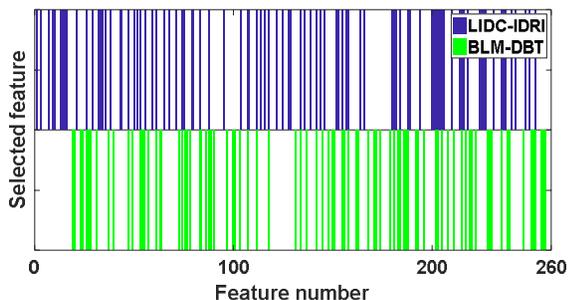

Fig. 6. Selected features for two datasets.

Fig. 7 shows the solutions generated, marked in blue, and the selected optimal solution, marked in red, for two running of the predictive model on LIDC-IDRI and BLM-DBT. The optimal solutions are always located at the "knee" point, which fully considers the trade-off between the two objective functions.

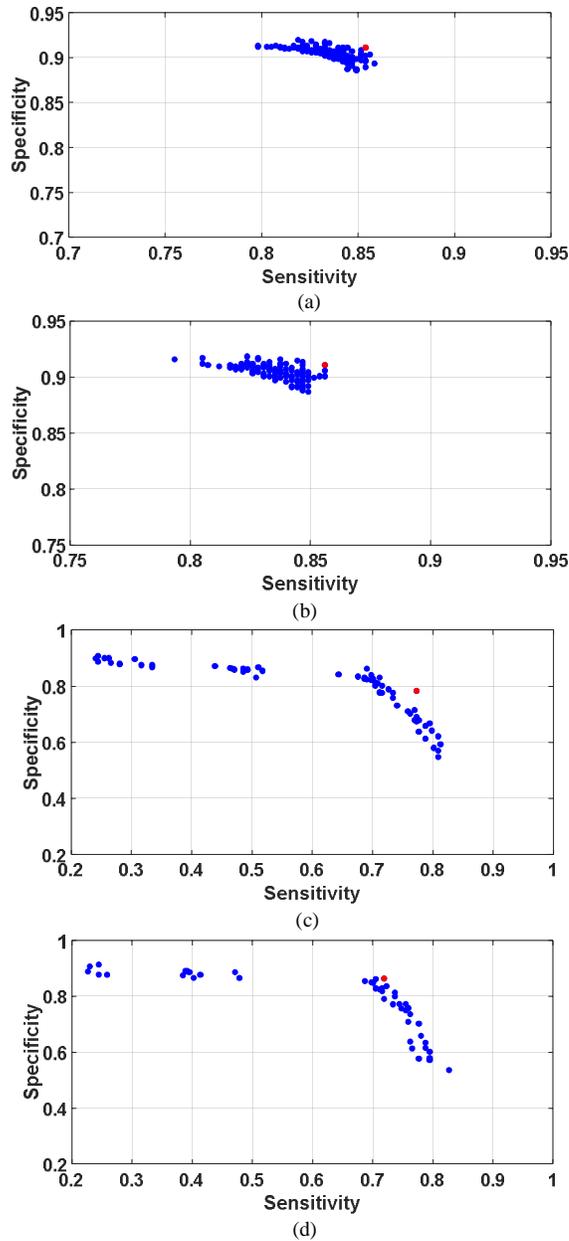

Fig. 7. Generated solutions (blue labels) and selected optimal solution (red label) for LIDC-IDRI (a-b) and BLM-DBT (c-d).

Fig. 8 shows the dissimilarity measure results for two running of the predictive model on the two datasets as the number of generation increases for the two datasets. With increasing of the generations, the dissimilarity measure results become smaller, showing the convergence of the algorithm. The proposed algorithm stopped automatically after a certain number of iterations, with different iterations for different datasets.

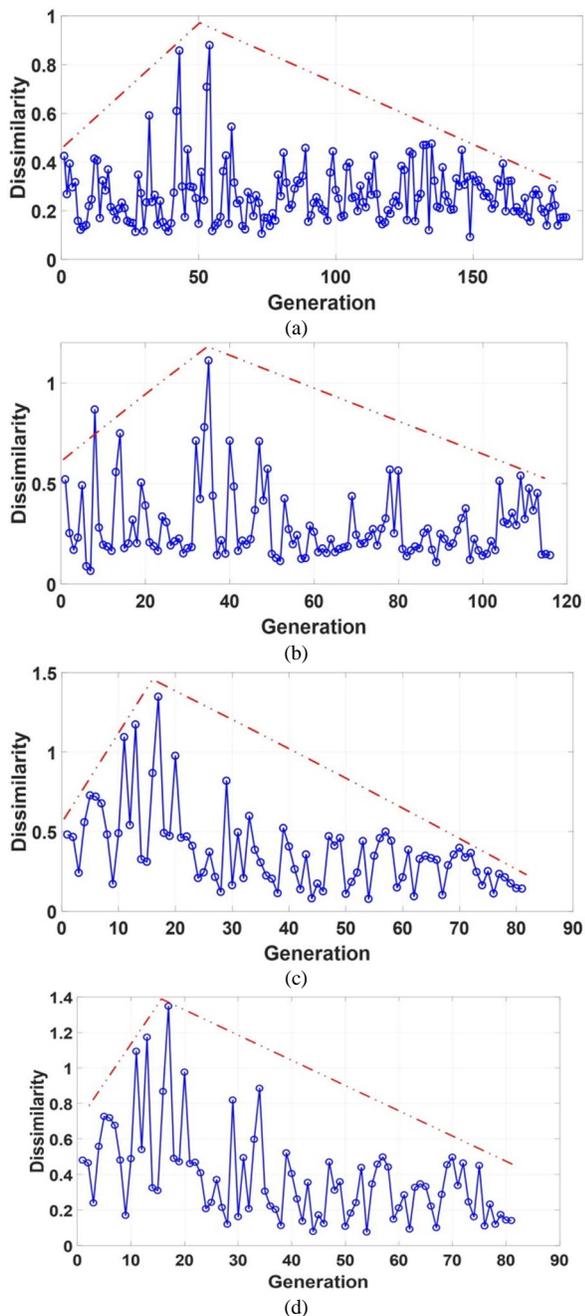

Fig. 8. Dissimilarity measure results with the increasing number of generations for LIDC-IDRI (a-b) and BLM-DBT (c-d). Blue lines are the dissimilarity measure results, and red lines are the dissimilarity trends.

To illustrate how the termination detection criterion affects the stability of the selected features, Fig. 8 shows the frequencies of the selected features in ten running of the predictive model on LIDC-IDRI and BLM-DBT. Compared with the entropy based termination criterion (ETC), METC selects the same features more frequently, showing better stability and repeatability.

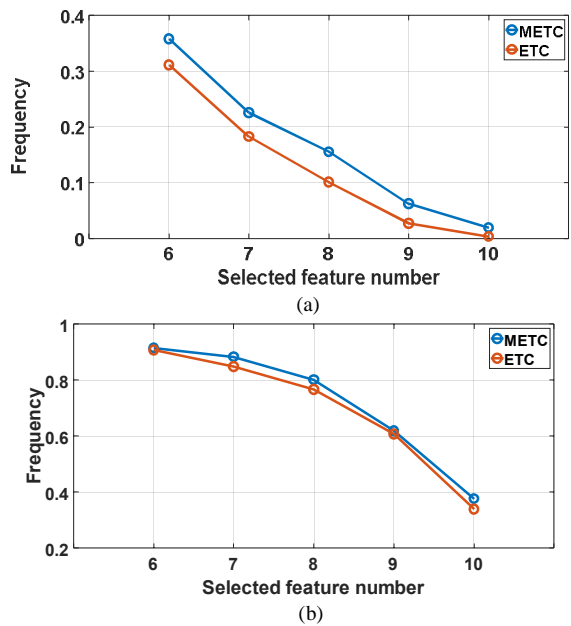

Fig. 9. The frequency of the selected features for ETC and METC: (a) LIDC-IDRI and (b) BLM-DBT.

### D. Sensitivity analysis of SMOLER

We investigated the influence of the weights and the reference number on the SMOLER selection results by analyzing the sensitivity. While analyzing the influence of the weight, the reference number r was set at 5. Assume that the four weights are denoted by $\{\omega_1, \omega_2, \omega_3, \omega_4\}$. We set $\omega_1 = \omega_2$ and $\omega_3 = \omega_4 = \frac{1-2\omega_1}{2}$, and $\omega_1$ increased from 0.25 to 0.4 in steps of 0.05. When we analyzed the sensitivity of the reference number, r increased from 5 to 11 in steps of 1, and the weight was fixed at {0.3, 0.3, 0.2, 0.2}. The sensitivity analysis of the weights for the two datasets is shown in Fig.10. The first row shows two examples for LIDC-IDRI, and the second row shows results for BLM-DBT. The results from LIDC-IDRI changed slightly with different weights, but there was no change for BLM-DBT. The results from analyzing the reference number (Fig. 11) show that there were no changes for any results, except in Fig. 11 (a), which shows a slight change (of 0.01) with different reference numbers. These results demonstrate the robustness of SMOLER.

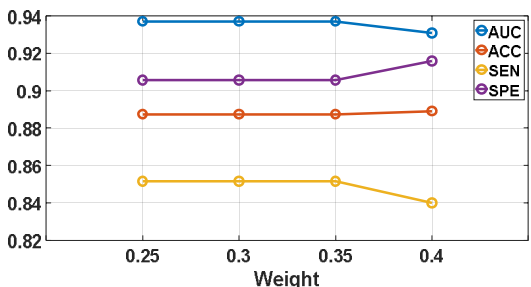

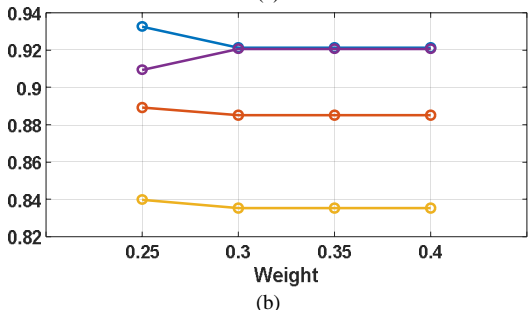

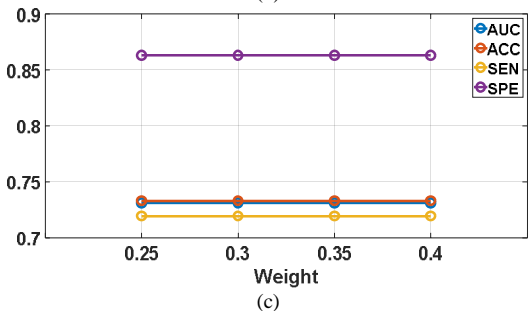

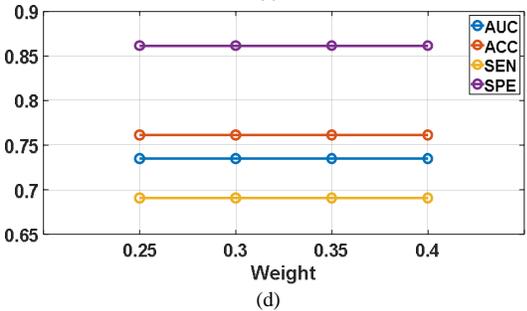

Fig. 10. Sensitivity analysis for weight in two running of the predictive model. The x-axis represents the increasing of $\omega_1$. The first row is the results for LIDC-IDRI, and the second row is the results for BLM-DBT.

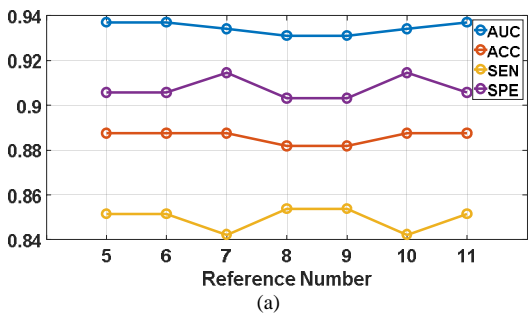

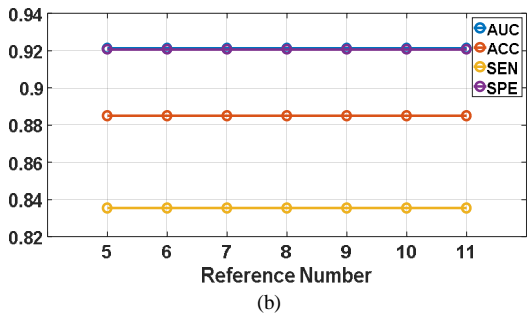

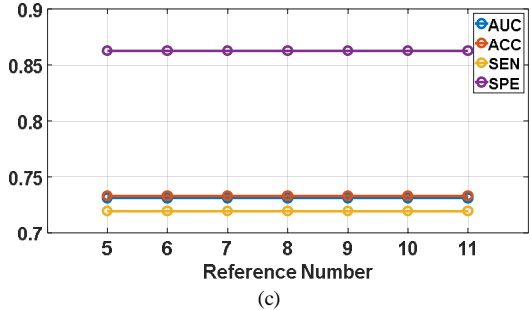

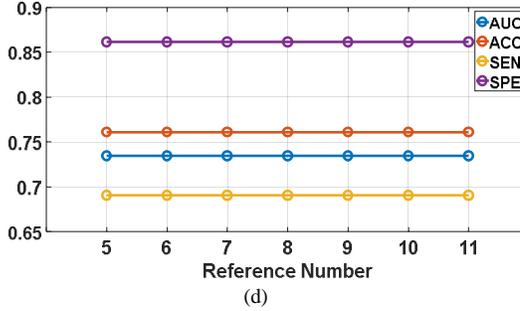

Fig. 11. Sensitivity analysis for reference number in two running of the predictive model. The first row is the results for LIDC-IDRI, and the second row is the results for BLM-DBT.

## IV. DISCUSSION AND CONCLUSION

Radiomics has achieved great success in many clinical applications recently. From a biomedical application perspective, a main contribution of radiomics is that it introduces quantitative imaging features into omics studies, which provides complementary information for omics data such as genomics, proteomics and supplies a new tool for personalized medicine. Meanwhile, radiomics is essentially a successful application of machine learning. Steps involved in a typical radiomics workflow include feature extraction, feature selection and model construction and the classical machine learning strategies are adopted in most radiomics based studies. Compared to conventional statistical approaches (which often focus on a limited number of variables or features), radiomics is considered as a more general approach as it can take a large number of features as input to build a predictive model. On the other hand, descriptive statistics can be integrated into radiomics, where different statistics tests such as correlational test, T-test, sign test, etc. can be used for filtering based feature selection in model construction [37].

Deep learning has also been applied successfully for many medical imaging processing and analysis tasks. In general, deep learning based strategies require a large scale dataset to obtain a good performance. However, it is often challenging to build a large dataset in many medical applications. Therefore, radiomics may be a preferred approach as it can adopt classification methods that are good at handling smaller scale data. Furthermore, since radiomics extracts features before building a predictive model, the meaning behind these features are known to an end user, which can make the model output to be interpreted in a more intuitive way. On the contrary, deep learning is typically implemented as an end-to-end black-box fashion, which makes it difficult to interpret the model output. Nevertheless, interpretable deep learning could make them more acceptable in clinical applications [38]. Additionally, the features learned by deep learning may contain complimentary information to the hand-crafted radiomic features, and the radiomics model performance may improve when combining learned features by deep learning [39].

This study proposed a multi-objective based feature selection (MO-FS) algorithm for selecting radiomic features for classification problems. In MO-FS, the algorithm's termination was detected automatically by a new modified entropy based termination criterion (METC) so that the generation does not need to be set manually. The METC also showed better stability for the selected feature subset than ETC. To select the optimal feature subset automatically, we developed a solution selection methodology for multi-objective learning using the evidential reasoning approach (SMOLER). Furthermore, we designed an adaptive mutation operation, which generates mutation probability automatically. We used two datasets to evaluate the performance of the MO-FS and compared it to several well-known feature selection methods. The classification performance showed that MO-FS outperformed other methods in selecting a feature set.

In this work, we mainly focus on binary classification problems. Therefore, the MO-FS takes sensitivity and specificity as the objectives. However, for multi-class prediction problems, such as tumor staging, more objectives (more than three) need to be considered. Thus, a many-objective based algorithm [40] needs to be developed to handle multi-class prediction. As more objectives and more rules are needed in SMOLER, manually setting the weights for different objectives becomes challenging, so a model for optimizing the weights is needed for SMOLER.

## ACKNOWLEDGEMENT

The authors would like to thank Dr. Jonathan Feinberg for editing the manuscript.

## REFERENCES

[1]  L. Wei, Y. Yang, R. M. Nishikawa, and Y. Jiang, "A study on several machine-learning methods for classification of malignant and benign clustered microcalcifications," *IEEE transactions on medical imaging*, vol. 24, pp. 371-380, 2005.

[2]  B. Verma, P. McLeod, and A. Klevansky, "Classification of benign and malignant patterns in digital mammograms for the diagnosis of breast cancer," *Expert systems with applications*, vol. 37, pp. 3344-3351, 2010.

[3]  U. Acharya, O. Faust, S. V. Sree, F. Molinari, R. Garberoglio, and J. Suri, "Cost-effective and non-invasive automated benign &

malignant thyroid lesion classification in 3D contrast-enhanced ultrasound using combination of wavelets and textures: a class of ThyroScan™ algorithms," *Technology in cancer research & treatment*, vol. 10, pp. 371-380, 2011.

[4]  H. J. Aerts, E. R. Velazquez, R. T. Leijenaar, C. Parmar, P. Grossmann, S. Cavalho, *et al.*, "Decoding tumour phenotype by noninvasive imaging using a quantitative radiomics approach," *Nature communications*, vol. 5, 2014.

[5]  P. Lambin, R. T. Leijenaar, T. M. Deist, J. Peerlings, E. E. de Jong, J. van Timmeren, *et al.*, "Radiomics: the bridge between medical imaging and personalized medicine," *Nature Reviews Clinical Oncology*, 2017.

[6]  J. Ma, Q. Wang, Y. Ren, H. Hu, and J. Zhao, "Automatic lung nodule classification with radiomics approach," in *Medical Imaging 2016: PACS and Imaging Informatics: Next Generation and Innovations*, 2016, p. 978906.

[7]  P. Cirujeda, Y. D. Cid, H. Müller, D. Rubin, T. A. Aguilera, B. W. Loo, *et al.*, "A 3-D Riesz-covariance texture model for prediction of nodule recurrence in lung CT," *IEEE transactions on medical imaging*, vol. 35, pp. 2620-2630, 2016.

[8]  H. Li, Y. Zhu, E. S. Burnside, E. Huang, K. Drukker, K. A. Hoadley, *et al.*, "Quantitative MRI radiomics in the prediction of molecular classifications of breast cancer subtypes in the TCGA/TCIA data set," *NPJ breast cancer*, vol. 2, p. 16012, 2016.

[9]  F. Valdora, N. Houssami, F. Rossi, M. Calabrese, and A. S. Tagliafico, "Rapid review: radiomics and breast cancer," *Breast cancer research and treatment*, pp. 1-13, 2018.

[10]  B. Xue, M. Zhang, W. N. Browne, and X. Yao, "A survey on evolutionary computation approaches to feature selection," *IEEE Transactions on Evolutionary Computation*, vol. 20, pp. 606-626, 2016.

[11]  G. Chandrashekar and F. Sahin, "A survey on feature selection methods," *Computers & Electrical Engineering*, vol. 40, pp. 16-28, 2014.

[12]  C. Parmar, P. Grossmann, J. Bussink, P. Lambin, and H. J. Aerts, "Machine learning methods for quantitative radiomic biomarkers," *Scientific reports*, vol. 5, 2015.

[13]  Z. Zhou, M. Folkert, P. Iyengar, K. Westover, Y. Zhang, H. Choy, *et al.*, "Multi-objective radiomics model for predicting distant failure in lung SBRT," *Physics in Medicine and Biology*, vol. 62, p. 4460, 2017.

[14]  J. García-Nieto, E. Alba, L. Jourdan, and E. Talbi, "Sensitivity and specificity based multiobjective approach for feature selection: Application to cancer diagnosis," *Information Processing Letters*, vol. 109, pp. 887-896, 2009.

[15]  H. Zhao, "A multi-objective genetic programming approach to developing Pareto optimal decision trees," *Decision Support Systems*, vol. 43, pp. 809-826, 2007.

[16]  C. Emmanouilidis, A. Hunter, and J. MacIntyre, "A multiobjective evolutionary setting for feature selection and a commonality-based crossover operator," in *Evolutionary Computation, 2000. Proceedings of the 2000 Congress on*, 2000, pp. 309-316.

[17]  L. S. Oliveira, R. Sabourin, F. Bortolozzi, and C. Y. Suen, "Feature selection using multi-objective genetic algorithms for handwritten digit recognition," in *Pattern Recognition, 2002. Proceedings. 16th International Conference on*, 2002, pp. 568-571.

[18]  B. Xue, M. Zhang, and W. N. Browne, "Particle swarm optimization for feature selection in classification: A multi-objective approach," *IEEE transactions on cybernetics*, vol. 43, pp. 1656-1671, 2013.

[19]  Y. Zhang, C. Xia, D. Gong, and X. Sun, "Multi-objective PSO algorithm for feature selection problems with unreliable data," in *International Conference in Swarm Intelligence*, 2014, pp. 386-393.

[20]  S. Vieira, J. Sousa, and T. A. Runkler, "Multi-criteria ant feature selection using fuzzy classifiers," *Swarm Intelligence for Multi-objective Problems in Data Mining*, vol. 242, pp. 19-36, 2009.

[21]  D. K. Saxena, A. Sinha, J. A. Duro, and Q. Zhang, "Entropy-Based Termination Criterion for Multiobjective Evolutionary Algorithms," *IEEE Transactions on Evolutionary Computation*, vol. 20, pp. 485-498, 2016.

[22]  E. Zio and R. Bazzo, "A comparison of methods for selecting preferred solutions in multiobjective decision making," *Computational intelligence systems in industrial engineering*, pp. 23-43, 2012.

[23] J.-B. Yang and M. G. Singh, "An evidential reasoning approach for multiple-attribute decision making with uncertainty," *IEEE Transactions on systems, Man, and Cybernetics,* vol. 24, pp. 1-18, 1994.

[24] J.-B. Yang and D.-L. Xu, "On the evidential reasoning algorithm for multiple attribute decision analysis under uncertainty," *Systems, Man and Cybernetics, Part A: Systems and Humans, IEEE Transactions on,* vol. 32, pp. 289-304, 2002.

[25] Z.-G. Zhou, F. Liu, L.-C. Jiao, Z.-L. Wang, X.-P. Zhang, X.-D. Wang*, et al.,* "An evidential reasoning based model for diagnosis of lymph node metastasis in gastric cancer," *BMC medical informatics and decision making,* vol. 13, p. 123, 2013.

[26] S. G. Armato, G. McLennan, L. Bidaut, M. F. McNitt-Gray, C. R. Meyer, A. P. Reeves*, et al.,* "The lung image database consortium (LIDC) and image database resource initiative (IDRI): a completed reference database of lung nodules on CT scans," *Medical physics,* vol. 38, pp. 915-931, 2011.

[27] M. Gong, L. Jiao, H. Du, and L. Bo, "Multiobjective immune algorithm with nondominated neighbor-based selection," *Evolutionary Computation,* vol. 16, pp. 225-255, 2008.

[28] M. G. Kendall, A. Stuart, and J. K. Ord, *The advanced theory of statistics* vol. 1: JSTOR, 1948.

[29] J.-B. Yang, "Rule and utility based evidential reasoning approach for multiattribute decision analysis under uncertainties," *European Journal of Operational Research,* vol. 131, pp. 31-61, 2001.

[30] Y.-M. Wang, J.-B. Yang, and D.-L. Xu, "Environmental impact assessment using the evidential reasoning approach," *European Journal of Operational Research,* vol. 174, pp. 1885-1913, 2006.

[31] K. Deb, A. Pratap, S. Agarwal, and T. Meyarivan, "A fast and elitist multiobjective genetic algorithm: NSGA-II," *IEEE transactions on evolutionary computation,* vol. 6, pp. 182-197, 2002.

[32] Q. Zhang and H. Li, "MOEA/D: A multiobjective evolutionary algorithm based on decomposition," *IEEE Transactions on evolutionary computation,* vol. 11, pp. 712-731, 2007.

[33] G. Roffo, S. Melzi, and M. Cristani, "Infinite feature selection," in *Proceedings of the IEEE International Conference on Computer Vision,* 2015, pp. 4202-4210.

[34] G. Roffo, S. Melzi, U. Castellani, and A. Vinciarelli, "Infinite Latent Feature Selection: A Probabilistic Latent Graph-Based Ranking Approach," *arXiv preprint arXiv:1707.07538,* 2017.

[35] Z.-G. Zhou, F. Liu, L.-C. Jiao, L.-L. Li, X.-D. Wang, S.-P. Gou*, et al.,* "Object information based interactive segmentation for fatty tissue extraction," *Computers in biology and medicine,* vol. 43, pp. 1462-1470, 2013.

[36] A. Krizhevsky, I. Sutskever, and G. E. Hinton, "Imagenet classification with deep convolutional neural networks," in *Advances in neural information processing systems,* 2012, pp. 1097-1105.

[37] I. A. Gheyas and L. S. Smith, "Feature subset selection in large dimensionality domains," *Pattern recognition,* vol. 43, pp. 5-13, 2010.

[38] Q. Zhang, Y. Nian Wu, and S.-C. Zhu, "Interpretable convolutional neural networks," in *Proceedings of the IEEE Conference on Computer Vision and Pattern Recognition,* 2018, pp. 8827-8836.

[39] P. Yan, H. Guo, G. Wang, R. De Man, and M. K. Kalra, "Hybrid deep neural networks for all-cause Mortality Prediction from LDCT Images," *arXiv preprint arXiv:1810.08503,* 2018.

[40] H. Ishibuchi, N. Tsukamoto, and Y. Nojima, "Evolutionary many-objective optimization," in *Genetic and Evolving Systems, 2008. GEFS 2008. 3rd International Workshop on,* 2008, pp. 47-52.

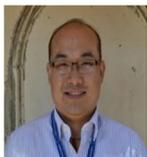

**Zhiguo Zhou** received the B.S. and Ph.D. degrees in Computer Science and Technology from Xidian University, Xi'an, China, in 2008 and 2014, respectively.

He joined the University of Texas Southwestern Medical Center, Dallas, TX, in 2014 as a Postdoctoral Researcher, where he is currently an instructor in Department of Radiation Oncology. He was a visiting scholar with Leiden University, Leiden, the Netherlands, from 2013 to 2014. He has published more than 20 peer reviewed journal papers. He is also the editorial board member of three international journals and reviewer of 18 journals. His current research interests include outcome prediction in cancer therapy, radiomics, medical image analysis, deep learning, machine learning and artificial intelligence.

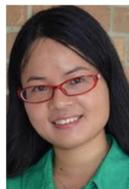

**Shulong Li** received the B. S. in Mathematics from Minnan Normal University, Zhangzhou, China, in 2003 and Ph. D degrees in Mathematics from Sun Yat-sen University, Guangzhou, China, 2008.

She joined the Southern Medical University, Guangzhou, China, in 2008 as a lecture, where she is currently an associate professor in School of Biomedical Engineering. She is a visiting assistant professor with University of Texas Southwestern Medical Center, Dallas, TX, from 2016 to 2017. She has published more than ten journal papers. Her current research interests include outcome prediction in cancer therapy, radiomics, medical image analysis, deep learning, machine learning and artificial intelligence.

**Genggeng Qin** is currently an associate chief physician in Department of Radiology, Nanfang Hospital, Southern Medical University. He

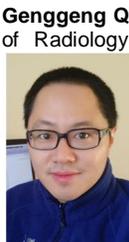

received the Bachelor of Medicine degrees from Sun Yat-sen University, Guangzhou, China, Master of Medicine degrees from Southern Medical University, Guangzhou, China. He is currently an visiting scholar in Department of Radiation Oncology the University of Texas Southwestern Medical Center, Dallas, TX. He has published 26 journal papers. His current research interests include radiomics, medical image analysis, deep learning, machine learning and artificial intelligence.

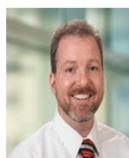

**Michael Folkert** received his Ph.D. in Radiological Sciences from the Massachusetts Institute of Technology in Cambridge, MA USA in 2005 and M.D. from Harvard Medical School in 2009.

He joined the University of Texas Southwestern Medical Center, Dallas, TX, in 2014 as an Assistant Professor in the Department of Radiation Oncology, and serves as the Medical Residency Director, Director of the Intraoperative Radiation Therapy program, and Co-Director of the Brachytherapy program. He has published 37 journal papers and 6 book chapters. His current research interests include outcome prediction in cancer therapy, radiomics/medical image analysis, brachytherapy applicator development, and interventional techniques for toxicity reduction. His clinical interests include management of ocular, gastrointestinal, genitourinary, musculoskeletal, and spine malignancies.

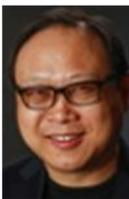

**Steve Jiang** received his B.S. degree in Theoretical Physics and M.A. degree in Physics from Sichuan University, Chengdu, China in 1990 and 1993, respectively, and Ph.D. degree in Medical Physics from Medical college of Ohio-Toledo in 1998.

He is currently a full professor, director in division of Medical Physics and Engineering, vice chair in Department of Radiation Oncology, at the University of Texas Southwestern Medical Center. Dr. Jiang has published more than 100 peer review journal papers. His research interests include artificial intelligence in medicine, GPU and cloud based automated treatment planning, and On-line re-planning for adaptive radiotherapy.

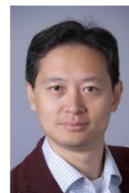

**Jing Wang** received his B.S. degree in Materials Physics from University of Science and Technology of China in 2001, M.A. and Ph.D. degrees in physics from the State University of New York at Stony Brook in 2003 and 2006, respectively. He finished his postdoctoral training in the Department of Radiation Oncology at Stanford University in 2009.

He is currently an Associate Professor and Medical Physicist in the Department of Radiation Oncology at the University of Texas Southwestern Medical Center. Dr. Wang has published more than 80 peer reviewed journal papers. His research focuses on medical image reconstruction/processing, machine learning and its applications in radiation therapy.